\documentclass[nolinenumbers]{iucrjournals}
\usepackage{times}  % DO NOT CHANGE THIS
\usepackage{helvet}  % DO NOT CHANGE THIS
\usepackage{courier}  % DO NOT CHANGE THIS
\usepackage{url}  % DO NOT CHANGE THIS
\usepackage{graphicx} % DO NOT CHANGE THIS
\usepackage{caption} % DO NOT CHANGE THIS AND DO NOT ADD ANY OPTIONS TO IT
\usepackage{algorithm}
\usepackage{algorithmic}
\usepackage{amsthm}
\usepackage{amssymb}
\usepackage{amsmath}
\usepackage{booktabs}   
\usepackage{amsmath,amsfonts,bm}

\def\eqref#1{equation~\ref{#1}}
\def\1{\bm{1}}

\DeclareMathAlphabet{\mathsfit}{\encodingdefault}{\sfdefault}{m}{sl}
\SetMathAlphabet{\mathsfit}{bold}{\encodingdefault}{\sfdefault}{bx}{n}

\usepackage{amssymb}
\usepackage{wrapfig}
\usepackage{mathtools}
\usepackage{tcolorbox}
\usepackage{subcaption}
\usepackage{comment}
 \usepackage{graphicx} 

\usepackage{newfloat}
\usepackage{listings}
\DeclareCaptionStyle{ruled}{labelfont=normalfont,labelsep=colon,strut=off} % DO NOT CHANGE THIS
\floatstyle{ruled}
\newfloat{listing}{tb}{lst}{}
\floatname{listing}{Listing}
\theoremstyle{plain}

\theoremstyle{definition}

\theoremstyle{remark}

\title{Reparameterization Flow Policy Optimization}

\author{Hai Zhong}%
\author{Zhuoran Li}%
\author{Xun Wang}%
\author{Longbo Huang \thanks{{ Corresponding Author}}}

\affil{Institute for Interdisciplinary Information Sciences (IIIS), Tsinghua University \{zhongh22,lizr20,wang-x24\}@mails.tsinghua.edu.cn, longbohuang@tsinghua.edu.cn }

\begin{document}

\maketitle

\begin{abstract}
Reparameterization Policy Gradient (RPG) has emerged as a powerful paradigm for model-based reinforcement learning, enabling high sample efficiency by backpropagating gradients through differentiable dynamics. However, prior RPG approaches have been predominantly restricted to Gaussian policies, limiting their performance and failing to leverage recent advances in generative models. In this work, we identify that flow policies, which generate actions via differentiable ODE integration, naturally align with the RPG framework, a connection not established in prior work. However, naively exploiting this synergy proves ineffective, often suffering from training instability and a lack of exploration. We propose Reparameterization Flow Policy Optimization (RFO). RFO computes policy gradients by backpropagating jointly through the flow generation process and system dynamics, unlocking high sample efficiency without requiring intractable log-likelihood calculations. RFO includes two tailored regularization terms for stability and exploration. We also propose a variant of RFO with action chunking. Extensive experiments on diverse locomotion and manipulation tasks, involving both rigid and soft bodies with state or visual inputs, demonstrate the effectiveness of RFO. Notably, on a challenging locomotion task controlling a soft-body quadruped, RFO achieves almost $2\times$ the reward of the state-of-the-art baseline.
\end{abstract}

\section{Introduction}
Reparameterization Policy Gradient (RPG)~\cite{SHAC,MonteCarlogradientestimationinmachinelearning,gao2024adaptivegradient,RPO} is a model-based paradigm that reparameterizes the policy's action distribution as a differentiable transformation from a source distribution~\cite{kingma:vae,rezende:vae}. By directly backpropagating gradients through the trajectory, RPG typically exhibits lower variance than likelihood-ratio estimators such as REINFORCE~\cite{Reinforce}, thereby enabling significantly higher sample efficiency~\cite{MonteCarlogradientestimationinmachinelearning,SHAC,AHAC}. This efficiency has driven recent successes in robotic applications leveraging differentiable simulators, ranging from quadrupeds to quadrotors~\cite{ResiudalDiff,NMIFLY,DIFFFLYVISUAL,learningdepolyablelocomotion,learningonthefly}. 

However, prior RPG approaches have predominantly focused on Gaussian policies~\cite{SHAC,SAPO,RPO}, limiting their expressiveness in modeling complex action distributions. In contrast, flow policies trained via imitation learning have recently demonstrated remarkable success in generating robotic actions~\cite{pi05,pi0,black2025realtime,Flowmatching}, owing to their superior expressiveness, implementation simplicity, and fast inference. To overcome the limitations of demonstrations, there has been a surge of research interest in leveraging reinforcement learning (RL) to enable flow policies to learn directly from their own experience~\cite{pi06,FLowmatchingPolicyGradients,SACFLOW,reinflow,GORL,FuchunFlow}.

In this work, we identify the key insight that flow policies inherently parameterize a differentiable mapping from source noise to actions, thus naturally aligning with the RPG framework. Leveraging this synergy, we view the flow generation process as an integral part of the rollout trajectory, enabling policy gradients to be computed by backpropagating directly through both the ODE integration and system dynamics—a novel connection not established in prior work. However, directly optimizing flow policies with RPG fails to yield strong performance, primarily due to optimization instability and insufficient exploration. 

Hence, we propose \textbf{Reparameterization Flow Policy Optimization (RFO)}, an on-policy RL algorithm optimizing flow policies with RPG. Beyond the core RPG mechanism, we introduce two regularization terms crucial for stable training. The first leverages actions sampled during rollouts as targets for Conditional Flow Matching (CFM) to stabilize policy updates, while the second utilizes targets from the uniform distribution to explicitly encourage exploration. Additionally, we propose a variant of RFO incorporating an action chunking mechanism~\cite{pi05,black2025realtime,RLactionChunk} to improve the temporal consistency of generated actions. We evaluate our method across a diverse suite of locomotion and manipulation tasks, involving both rigid and soft bodies, using state or visual inputs. Notably, RFO achieves nearly $2\times$ the reward of the state-of-the-art (SOTA) baseline on a challenging soft-body quadruped locomotion task with a high-dimensional action space ($\mathbb{R}^{222}$).

RFO represents a significant contribution to both the flow-based RL and RPG communities. From the perspective of flow-based RL, RFO uniquely combines the strengths of distinct paradigms: it inherits the high sample efficiency of RPG while eliminating the need for intractable action log-likelihood computations—benefits typically associated with off-policy methods. Simultaneously, as an on-policy approach, RFO seamlessly integrates with massively parallel differentiable physics simulators, such as Rewarped, DFlex, and Newton~\cite{SAPO,SHAC,AHAC,IssacLab}, to significantly accelerate training. From the RPG perspective, to the best of our knowledge, RFO is the first work to successfully investigate and demonstrate the effectiveness of incorporating state-of-the-art generative flow models. 

To summarize, our contributions are threefold: (i) We propose a novel paradigm for training flow policies using Reparameterization Policy Gradients, unlocking high sample efficiency without approximating intractable likelihoods. (ii) We introduce Reparameterization Flow Policy Optimization (RFO), an algorithm equipped with novel regularization mechanisms for stability and exploration, further extended with an action-chunking formulation. (iii) We conduct extensive experiments on differentiable simulators, demonstrating that RFO consistently achieves strong performance across challenging rigid and soft-body tasks compared to existing baselines.

\section{Related Work}

\subsection{Online Reinforcement Learning Algorithms for Flow and Diffusion Policy}
Online RL algorithms for flow and diffusion policies can be categorized as off-policy or on-policy. Off-policy algorithms rely on Q-functions to train generative policies~\cite{LearningadiffusionmodelpolicyfromrewardsviaQ-scorematching,DACERV2,DIME,FuchunFlow,DPMD,FPMD,MaxEntDP,DIPO,SACFLOW,QVPO,DRAC,FQL}, offering the advantage of bypassing the calculation of action log-likelihoods. One approach for policy optimization involves backpropagating Q-gradients to the noise prediction or vector field network~\cite{DACER,DACERV2,FuchunFlow,DRAC,SACFLOW,DIME}. This is achieved by reparameterizing the flow or diffusion policy as a transformation from sampled noise to actions, which allows for the direct backpropagation of Q-function gradients through this differentiable transformation to the policy network. Another class of approaches exploits the intrinsic connection between the score function (or velocity field) and Q-functions~\cite{LearningadiffusionmodelpolicyfromrewardsviaQ-scorematching,MaxEntDP,Iterateddenoisingenergymatching,QVPO}. For instance, \cite{LearningadiffusionmodelpolicyfromrewardsviaQ-scorematching} proposes matching the score functions to Q-function gradients. Meanwhile, \cite{MaxEntDP,DPMD,QVPO} utilize a Q-weighted regression loss to train the noise prediction network, whereas \cite{FPMD} trains the velocity network via a Q-weighted conditional flow matching loss. 

On-policy RL algorithms for flow and diffusion policies~\cite{GENPO,DPPO,reinflow,NCDPO,FLowmatchingPolicyGradients} must address the challenge of approximating otherwise intractable action log-likelihoods. One prevalent approach is to treat the sampled noise as an augmented state, which enables the calculation of action likelihoods conditioned on the noise. For instance, DPPO~\cite{DPPO} constructs a two-level Markov Decision Process that propagates policy gradient signals to each denoising step. Similarly, NCDPO and Reinflow~\cite{NCDPO,reinflow} treat the action generation as a deterministic process by conditioning on all sampled noise, and additionally train a noise injection network to compute action log-likelihoods. Alternatively, FPO~\cite{FLowmatchingPolicyGradients} approximates action log-likelihoods via the conditional flow matching loss. GenPO~\cite{GENPO} introduces a specialized doubled dummy action generation mechanism, which enables the calculation of exact action log-likelihoods. In contrast, RFO entirely bypasses likelihood computation by leveraging reparameterization gradients through differentiable dynamics, which typically yields lower-variance gradient estimates than REINFORCE-like policy gradients~\cite{MonteCarlogradientestimationinmachinelearning}.

\subsection{Reparameterization Policy Gradient}
Reparameterization Policy Gradient exploits the reparameterization trick~\cite{kingma:vae,rezende:vae,MonteCarlogradientestimationinmachinelearning} to render the action sampling process differentiable. By defining the action as a deterministic transformation of a noise variable drawn from a fixed source distribution, RPG enables the computation of low-variance policy gradients by backpropagating directly through the system dynamics. These dynamics are typically provided by either differentiable simulators~\cite{SHAC,DiffTaichi,SAPO} or learned world models~\cite{DreamerV1,SVG,roboticworldmodel}. Consequently, RPG generally yields gradient estimates with significantly lower variance compared to likelihood-ratio estimators like REINFORCE~\cite{MonteCarlogradientestimationinmachinelearning,Sutton:Reinforce}.  Prominent RPG algorithms, such as SHAC and AHAC~\cite{SHAC,AHAC}, further stabilize training by truncating the backpropagation horizon and leveraging value function gradients to estimate long-term returns. While prior work has investigated the use of multimodal policies within RPG~\cite{RPGmultimodal}, the integration of modern expressive generative models—specifically Diffusion and Flow Matching—remains largely unexplored. RFO bridges this gap by introducing the first RPG-based framework tailored for Flow Matching policies.

\begin{figure*}
    \centering
    \vspace{-5pt}
    \includegraphics[width=0.8\linewidth]{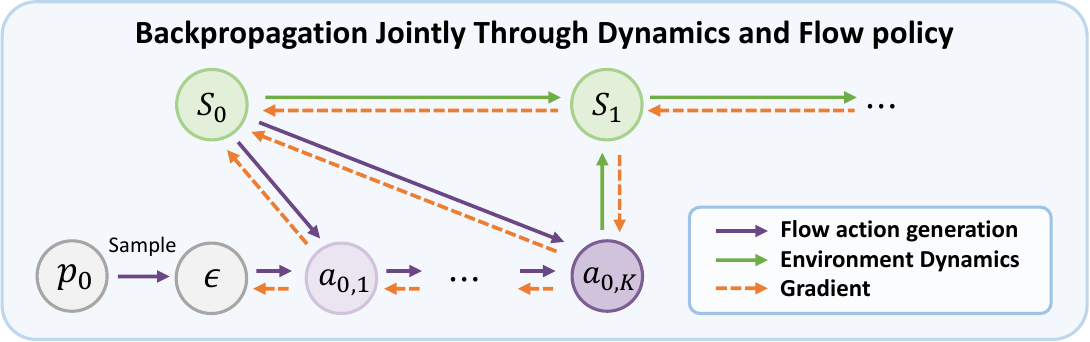}
    \caption{RFO optimizes the flow policy by jointly backpropagating through the dynamics and the flow policy.}
    \label{fig: BPTTgraph_a}  
\end{figure*}

\section{Preliminaries}
\subsection{Reinforcement Learning Problem Formulation}
We formulate the reinforcement learning (RL) problem as a Markov Decision Process (MDP)~\cite{RLbook}, defined by the tuple $(\mathcal{S}, \mathcal{A}, p, r, \rho_0, \gamma)$. Here, $\mathcal{S}$ denotes the state space, and $\mathcal{A}$ denotes the action space. The transition function $p(s'|s, a)$ defines the probability density of transitioning to the next state $s' \in \mathcal{S}$ given the current state $s \in \mathcal{S}$ and action $a \in \mathcal{A}$. The term $r(s,a)$ represents the reward function, $\rho_0$ denotes the initial state distribution, and $\gamma \in [0, 1)$ is the discount factor. In this work, we consider a continuous and \textbf{bounded} action space $\mathcal{A}$. The objective of RL is to maximize the expected discounted cumulative reward:
\begin{equation}
    J(\theta) = \mathbb{E}_{\tau \sim \pi_{\theta}} \left[ \sum_{t=0}^{\infty} \gamma^t r(s_t, a_t) \right],
\end{equation}
where $\tau = (s_0, a_0, s_1, a_1, \dots)$ denotes a trajectory sampled from the distribution induced by the policy $\pi_{\theta}$, with $s_0 \sim \rho_0$.

\subsection{Flow Model and Flow Policy}
\paragraph{Flow Models and Flow Matching.} Flow models~\cite{Flowmatching,NeuralODE,NormalizingFLowWithStochasticInterpolants,RecitifiedFLOW} constitute a class of generative models that transform samples from a source distribution $x_0 \sim p_0$ (e.g., standard Gaussian) to a target distribution $x_1 \sim p_1$ via an invertible flow map $\phi: [0,1] \times \mathbb{R}^d \rightarrow \mathbb{R}^d$. This flow map is induced by a time-dependent, neural-network-parameterized vector field $v_{\theta}(x,u)$~\cite{NeuralODE}, defined by the ordinary differential equation (ODE):
\begin{equation}
    \frac{d}{du} \phi(u,x_0) = v_{\theta}(\phi(u,x_0),u),
\end{equation}
where $\phi(0, x_0) = x_0$ and $u$ denotes the flow time. To learn the parameterized vector field $v_{\theta}$, the Conditional Flow Matching (CFM) objective is proposed~\cite{Flowmatching}. Specifically, CFM constructs a linear interpolation path conditioned on the source noise $x_0$ and the target sample $x_1$, where the intermediate state is given by:
\begin{equation}
    x_u= (1-u)x_0 + u x_1, \quad u \in [0, 1].
\end{equation}
Consequently, the CFM objective is defined as:
\begin{equation}
\begin{aligned}
    & \mathcal{L}_{\text{CFM}}(\theta) \\
    & = \mathbb{E}_{u \sim \mathcal{U}[0,1], x_0 \sim p_0, x_1 \sim p_1} \Big[ \left\| v_{\theta}(x_u, u) - (x_1 - x_0) \right\|^2 \Big],
\end{aligned}
\end{equation}
where $\mathcal{U}[0,1]$ is the uniform distribution. 
\paragraph{Flow Policy.} Flow models serve as a powerful policy class for generating actions~\cite{pi05,pi06,black2025realtime} from potentially complex, multi-modal distributions. To define a state-conditioned policy, the vector field $v_{\theta}(x, u| s)$ is conditioned on the current state $s \in \mathcal{S}$. An action $a$ is generated by first sampling $x_0$ from the source distribution $p_0$, and subsequently solving the following ODE from $u=0$ to $u=1$:
\begin{equation} \label{eq:FLOWODE}
    \frac{dx}{du} = v_{\theta}(x, u | s), \quad \text{with } x(0) = x_0.
\end{equation}
The core challenge in utilizing flow policies for RL lies in propagating reward signals to optimize the vector field $v_{\theta}$.

\subsection{Reparameterization Policy Gradient}
The Reparameterization Policy Gradient method builds upon the reparameterization trick~\cite{kingma:vae,rezende:vae}. Instead of directly parameterizing a potentially complex, multi-modal action distribution, RPG parameterizes the transformation from a source distribution to the action distribution. Specifically, given noise $\epsilon$ sampled from a fixed source distribution, the action $a$ is computed via a transformation function $a = f_{\theta}(\epsilon; s)$ (typically parameterized by a neural network), conditioned on the state $s$. Prior RPG-based approaches have predominantly focused on Gaussian policies~\cite{SHAC,AHAC,SAPO,gao2024adaptivegradient,RPO}, effectively modeling the transformation from a standard Gaussian distribution to a state-dependent Gaussian distribution.

Following previous RPG work~\cite{SHAC,AHAC,SAPO}, we focus on deterministic environment dynamics $s_{t+1}=g(s_t,a_t)$ and assume environment dynamics and the reward functions are differentiable.

Leveraging the reparameterization transformation, RPG computes the policy gradient by backpropagating through the environment dynamics, utilizing the dynamics Jacobians $\frac{\partial s_{t+1}}{\partial a_t}$ and $\frac{\partial s_{t+1}}{\partial s_t}$. The objective function and its gradient are defined as:
\begin{align}
    J(\theta) &= \mathbb{E}_{s_0, \epsilon_0, \epsilon_1, \dots} \left[ R(\tau)\right], \\
    \nabla_{\theta} J(\theta) &= \mathbb{E}_{s_0, \epsilon_0, \epsilon_1, \dots} \left[ \nabla_{\theta} R(\tau)\right],
\end{align}
where $R(\tau)= \sum_{t=0}^{\infty} \gamma^t r(s_t, a_t)$ denotes the cumulative discounted reward for the trajectory $\tau$. Crucially, this expectation is taken with respect to the initial state distribution and the sequence of independent noise variables sampled from the source distribution at each time step.

\section{Reparameterization Flow Policy Optimization}
In this section, we detail the proposed algorithm, Reparameterization Flow Policy. RFO comprises three key components. First, we derive the optimization of flow policies with RPG. Second, we introduce two regularization terms that are critical for RFO's performance. Finally, we present a variant of RFO that incorporates an action chunking mechanism.
\subsection{Reparameterization Policy Gradient for Flow Policy}
We now detail the first key component of RFO: reparameterization policy gradient for flow policies. We adopt the \textit{discretize-then-optimize} paradigm~\cite{DiscreteOptimizeVsOptimizeDiscrete} for flow policy optimization, consistent with prior studies on reinforcement learning with flow policies~\cite{reinflow,FuchunFlow,SACFLOW,GENPO}. Specifically, given the neural-network-parameterized vector field $v_{\theta}$, we perform numerical integration on the ODE using Euler method~\cite{EulermethodReference}. We employ the double-subscript notation $a_{t,k}$, where the subscript $t$ denotes the MDP time step and $k$ indexes the internal discrete integration steps. The generation process is formalized as follows:
\begin{equation}
\begin{aligned}
    a_{t, 0} &= \epsilon_t, \quad \text{where } \epsilon_t \sim p_{0}, \\
    a_{t, k+1} &= a_{t, k} + \Delta u \cdot v_{\theta}(a_{t, k}, u_k | s_t), \quad k=0, \dots, K-1, \\
    a_t &= a_{t, K},
\end{aligned}
\end{equation}
where $\Delta u = 1/K$ is the step size and $u_k = k \cdot \Delta u$ denotes the k-th flow time. Consequently, Euler integration process explicitly defines a differentiable transformation from the noise $\epsilon_t$ to the action $a_t$. We denote this transformation as $a_t = F_{\theta}(\epsilon_t; s_t)$. Based on this, we have the following key insight to optimize RFO:

% 确保导言区有 \usepackage{tcolorbox}

\begin{tcolorbox}[
    colback=white,          % 背景改为白色 (更像 Paper)
    colframe=black,         % 边框改为纯黑
    boxrule=0.8pt,          % 边框稍微细一点
    arc=2pt,                % 圆角稍微小一点，或者设为 0pt (直角)
    left=4pt, right=4pt,    % 内部边距
    top=4pt, bottom=4pt,
    boxsep=2pt
]
    \textbf{Key Insight:} Flow policies inherently parameterize a differentiable mapping from source noise to actions. This structure is naturally compatible with the Reparameterization Policy Gradient framework, enabling the computation of policy gradients via end-to-end backpropagation jointly through the environment dynamics and the flow policy.
\end{tcolorbox}

An illustration of this insight is shown in Figure~\ref{fig: BPTTgraph_a}. To formalize this insight, reparameterization policy gradient for the flow policy is expressed as:
\begin{equation} \label{eq:rpgFlow_full}
\begin{split}
    \nabla_{\theta} J(\theta) &= \mathbb{E}_{s_0,\epsilon_0,\epsilon_1,\dots} \left[ \nabla_{\theta} \sum_{t=0}^{\infty} \gamma^t  r\left(s_t, F_{\theta}(\epsilon_t; s_t)\right) \right],
\end{split}
\end{equation}
where $\epsilon_t$ denotes the noise sampled from the source distribution at time $t$, and $F_{\theta}$ represents the flow transformation. The gradient is computed via backpropagation through time (BPTT) on the computational graph, which includes both the flow ODE integration and the environment dynamics typically provided by a differentiable simulator or a learned world model.

In practice, backpropagating through long horizons can lead to exploding gradients. Therefore, we adopt the Short-Horizon Actor-Critic (SHAC) paradigm~\cite{SHAC,AHAC} by truncating the trajectory into short segments. We define the short-horizon surrogate objective $\hat{J}(\theta)$ as:
\begin{equation}
    \hat{J}(\theta) = \mathbb{E}_{s_0,\epsilon_{0},...} \left[ \sum_{t=0}^{H-1} \gamma^t r(s_{t}, F_{\theta}(\epsilon_t; s_t))) + \gamma^H V_{\omega}(s_{H}) \right].
\end{equation}
The gradient is thus \textit{approximated} as:
\begin{equation}
\label{eq:SHAC_rpg}
\begin{split}
    &\hat{\nabla}_{\theta} J(\theta) \\&= \mathbb{E} \bigg[ \nabla_{\theta} \bigg(  \sum_{t=0}^{H-1} \gamma^t r(s_{t}, F_{\theta}(\epsilon_t; s_t)) 
      + \gamma^H V_{\omega}(s_{H}) \bigg) \bigg],
\end{split}
\end{equation}
where $s_0$ represents the start state of the current segment, $V_{\omega}(s_H)$ is the terminal value estimate to account for future rewards.

\begin{figure}[t]
    \centering
    \begin{subfigure}[b]{0.45\columnwidth}
        \centering
        \includegraphics[width=\textwidth]{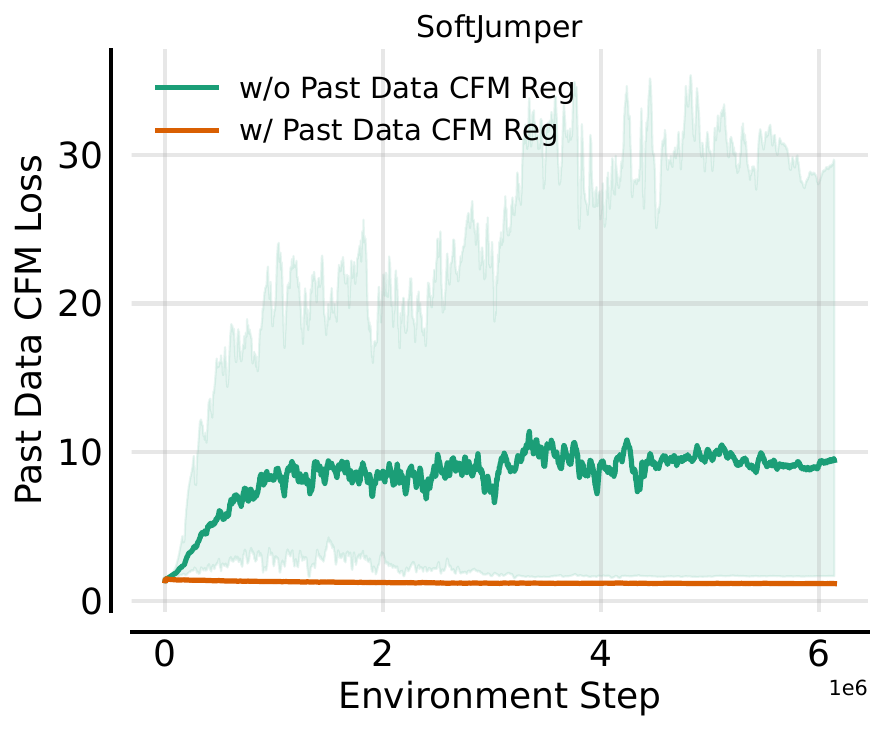}
        \caption{Past Data CFM Loss}
        \label{fig:cfm_loss}
    \end{subfigure}
    \hfill
    \begin{subfigure}[b]{0.45\columnwidth}
        \centering
        \includegraphics[width=\textwidth]{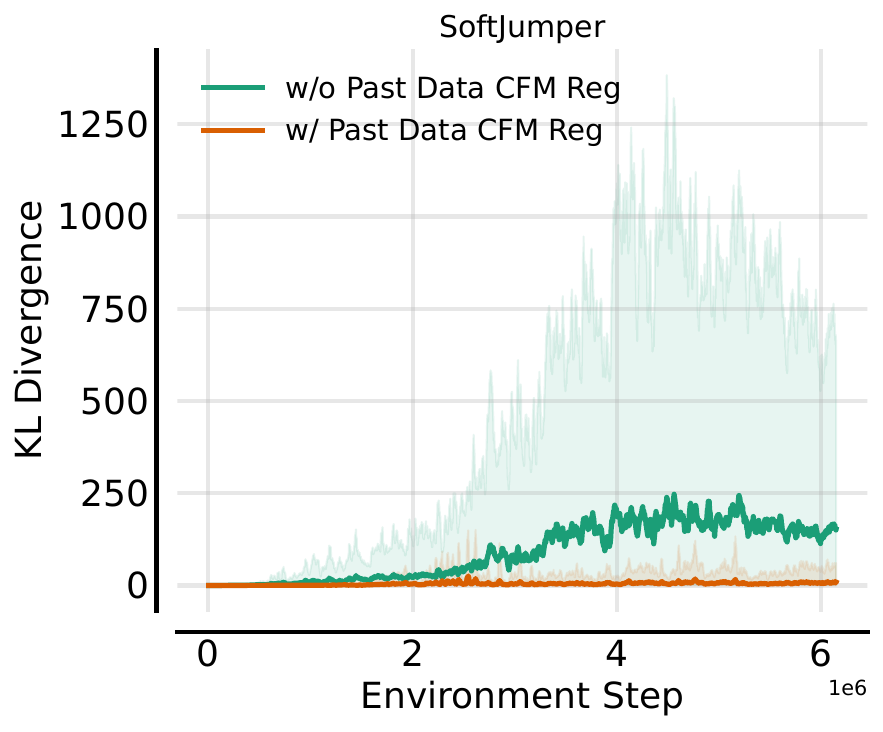}
        \caption{KL Divergence}
        \label{fig:kl_divergence}
    \end{subfigure}
\caption{Comparison of RFO training with and without past data CFM regularization on the Soft Jumper task. We monitored the Conditional Flow Matching (CFM) loss of the current policy evaluated on actions sampled from the immediately preceding iteration, as well as the KL divergence between consecutive policy updates.}
    \label{fig:cfm_kl_comparison}
\end{figure}

\subsection{Regularization Terms}
RFO incorporates two regularization terms that are crucial for performance. The first term addresses the stability of policy updates. RPG optimizes the policy by backpropagating gradients through the ODE integration steps, which can potentially disrupt the flow trajectories towards previously generated actions, rendering them unlikely to be re-sampled under the updated policy.

To illustrate this phenomenon, we conducted experiments on the Soft Jumper task using the Rewarped simulator~\cite{SAPO}. We monitored the Conditional Flow Matching (CFM) loss of the current policy evaluated on actions sampled from the immediately preceding iteration. As shown in Figure~\ref{fig:cfm_kl_comparison}, the CFM loss steadily increases and remains high, indicating that the updated vector field struggles to reconstruct past actions. Furthermore, we tracked the KL divergence between consecutive policy updates (KL divergence approximation details are provided in Appendix~\ref{appendix:KL}). The large KL divergence clearly signals training instability.

\begin{algorithm} [b]
\caption{Reparameterization Flow Policy Optimization (RFO)}
\label{alg:RFO}
\begin{algorithmic}[1]

\STATE Initialize $\theta, \omega$ and buffers $\mathcal{D}_{\text{recent}}$, $\mathcal{D}_{\text{rollout}}$.
\FOR{iteration $k = 1, 2, \dots$}
    \STATE Collect short-horizon trajectories with flow policy; update $\mathcal{D}_{\text{recent}}$ and $\mathcal{D}_{\text{rollout}}$.
    \STATE Compute RPG gradient for flow policy $\hat{\nabla} J$ via BPTT (Eq.~\eqref{eq:SHAC_rpg}).
    \STATE Compute regularization CFM loss gradients $\nabla \mathcal{L}_{\text{past}}$ (Eq.~\eqref{eq:cfmrollout}) and $\nabla \mathcal{L}_{\text{uni}}$ (Eq.~\eqref{eq:cfmuni}).
    \STATE Update flow policy with combined and weighted gradients (Eq.~\eqref{eq:total_loss}).
    \STATE Update critic $\omega$ for $L$ epochs by minimizing Eq.~\eqref{eq:value_loss}.
\ENDFOR
\end{algorithmic}
\end{algorithm}

Hence, intuitively, we aim to regularize the vector field to retain paths from the source distribution to these recently sampled actions. We propose the \textit{Past Data CFM Regularization}. This term treats the recent rollouts' actions as target samples and ensures the vector field points towards them. The objective $\mathcal{L}_{\text{past}}$ is defined as:
\begin{equation}
\begin{split}
    &\mathcal{L}_{\text{past}}(\theta)  \\
    &= \mathbb{E}_{\substack{u \sim \mathcal{U}[0,1], \epsilon \sim p_0, \\ (s, a) \sim \mathcal{D}_{\text{recent}}}}  \left[ \left\| v_{\theta}(\psi_{u}, u | s) - (a - \epsilon) \right\|^2 \right],
\end{split} \label{eq:cfmrollout}
\end{equation}
where $\mathcal{D}_{\text{recent}}$ is a buffer including state-action pairs from the two most recent iterations (including the current iteration's rollout data), and $u$ is the flow time. A discussion for the design of $\mathcal{D}_{\text{recent}}$ is in appendix~\ref{appendix:DesignChoice}. Here, $\psi_{u} = (1-u)\epsilon + u a$ represents the linear interpolation between the source noise $\epsilon$ and the target action $a$ at flow time $u$. The effectiveness of $\mathcal{L}_{\text{past}}$ is empirically validated in our ablation study in Section~\ref{sec:main_ablation}. While self-imitation strategies have appeared in diffusion RL contexts~\cite{NCDPO}, to the best of our knowledge, we are the first to uncover the distinct and fundamental role of Past Data CFM within the fundamentally different paradigm of RPG-driven flow policy optimization. Our analysis establishes that this regularization serves as a critical stabilizer for RPG-based flow optimization.

We introduce a second regularization term to explicitly promote exploration. Unlike prior Maximum Entropy approaches~\cite{MaxEntDP,DACER,DIME}, which rely on approximating the entropy of the generative policy, we adopt a more direct strategy. Given the bounded action space in our setting, we leverage the property that maximizing entropy is equivalent to minimizing the KL divergence from a uniform distribution (see Example 12.2.4 in \textit{Elements of Information Theory}~\cite{elementsofInformationTheory}). Motivated by this, we propose to guide the flow policy toward uniformly sampled target actions via the CFM objective. While QVPO~\cite{QVPO} explores a similar uniform-target regularization for diffusion policies, it employs state-dependent weighting. In contrast, we adopt a simple yet effective approach that treats all states encountered during rollouts equally. The proposed uniform exploration CFM objective, denoted as $\mathcal{L}_{\text{uni}}$, is defined as:
\begin{equation}
\begin{split}
    &\mathcal{L}_{\text{uni}}(\theta)  \\
    &= \mathbb{E}_{\substack{u \sim \mathcal{U}[0,1], \epsilon \sim p_0, \\ s \sim \mathcal{D}_{\text{rollout}}, a \sim p_{\text{uni}}}}  \left[ \left\| v_{\theta}(\psi_{u}, u | s) - (a - \epsilon) \right\|^2 \right],
\end{split} \label{eq:cfmuni}
\end{equation}
where $\mathcal{D}_{\text{rollout}}$ denotes the set of states visited during the current iteration's rollout, and $p_{\text{uni}}$ represents the uniform distribution over the bounded action space. As before, $\psi_{u}$ denotes the linear interpolation. The empirical effectiveness of $\mathcal{L}_{\text{uni}}$ is analyzed in Section~\ref{sec:main_ablation}.

\subsection{Value Function Training}

We optimize the critic by minimizing the standard mean squared error loss \cite{SAPO,SHAC}:
\begin{equation}
    \mathcal{L}(\omega) = \mathbb{E}\left[ \| V_{\omega}(s) - y(s) \|^2 \right],
    \label{eq:value_loss}
\end{equation}
where $y(s)$ denotes the TD-$\lambda$ target \cite{RLbook}. Following the dual-critic architecture in SAPO \cite{SAPO}, we calculate the target value by averaging the predictions of two separate critic networks.

\subsection{Overall Policy Objective and RFO Algorithm}

Combining the RPG objective for the flow policy (Eq.~\eqref{eq:SHAC_rpg}) with the  past data CFM regularization (Eq.~\eqref{eq:cfmrollout}) and uniform exploration CFM regularization (Eq.~\eqref{eq:cfmuni}), RFO minimizes the total policy loss:
\begin{equation}
    \mathcal{L}_{\text{policy}}(\theta) = - \hat{J}(\theta) + c_{\text{past}} \mathcal{L}_{\text{past}}(\theta) + c_{\text{uni}} \mathcal{L}_{\text{uni}}(\theta),
    \label{eq:total_loss}
\end{equation}
where $c_{\text{past}}$ and $c_{\text{uni}}$ are hyperparameters weighting the stability and exploration regularization terms, respectively.

The overall RFO algorithm is described in Algorithm~\ref{alg:RFO}. We follow the SHAC paradigm, rolling out with flow policies to collect short-horizon trajectories. Then, we compute policy gradients and the gradients of the two CFM regularization terms. Combining and weighting these gradients, we update the policy. Afterwards, we update the value function by minimizing Eq.~\eqref{eq:value_loss}.

\begin{figure*}[t]

    \centering
    % --- 第一行：四张图 ---
    \includegraphics[width=0.24\textwidth]{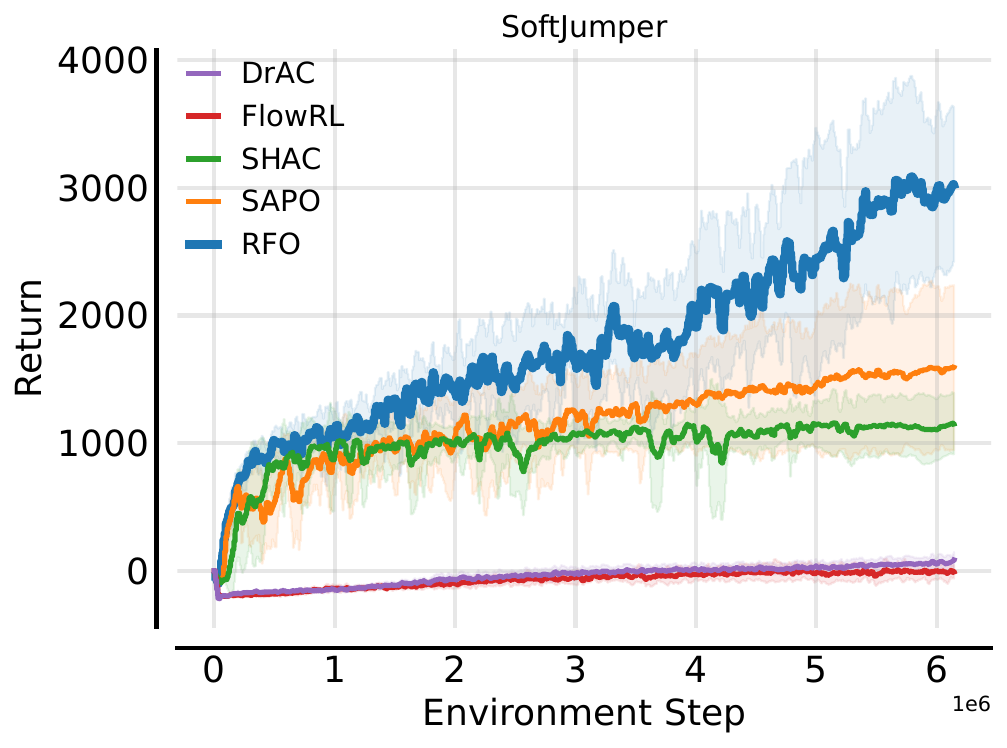}
    \hfill
    \includegraphics[width=0.24\textwidth]{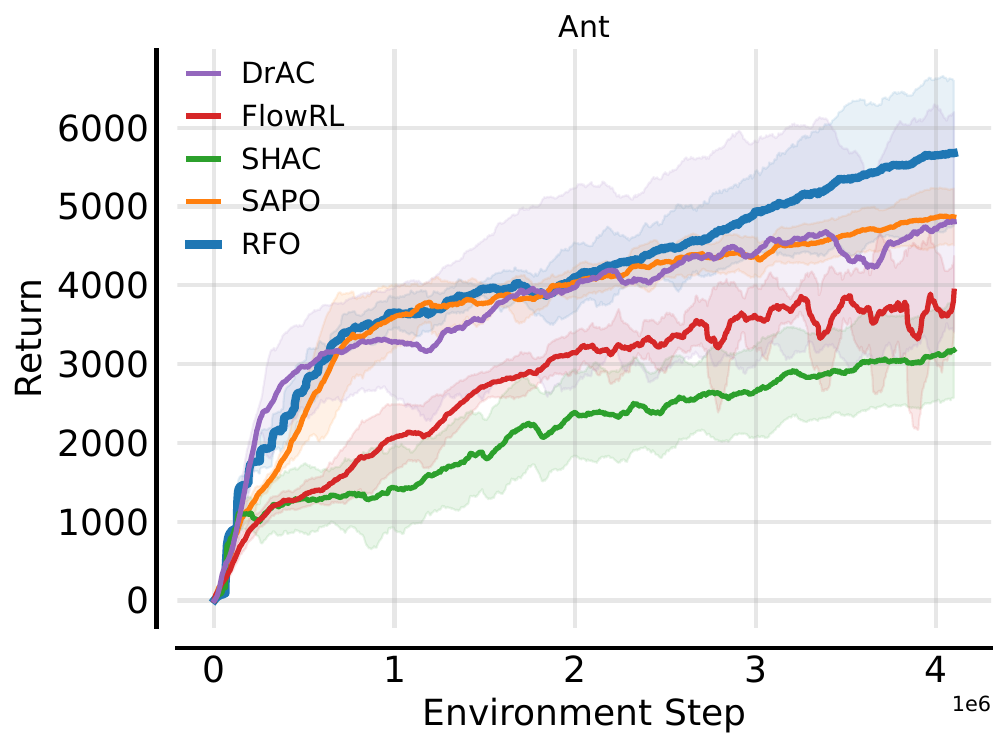}
    \hfill
    \includegraphics[width=0.24\textwidth]{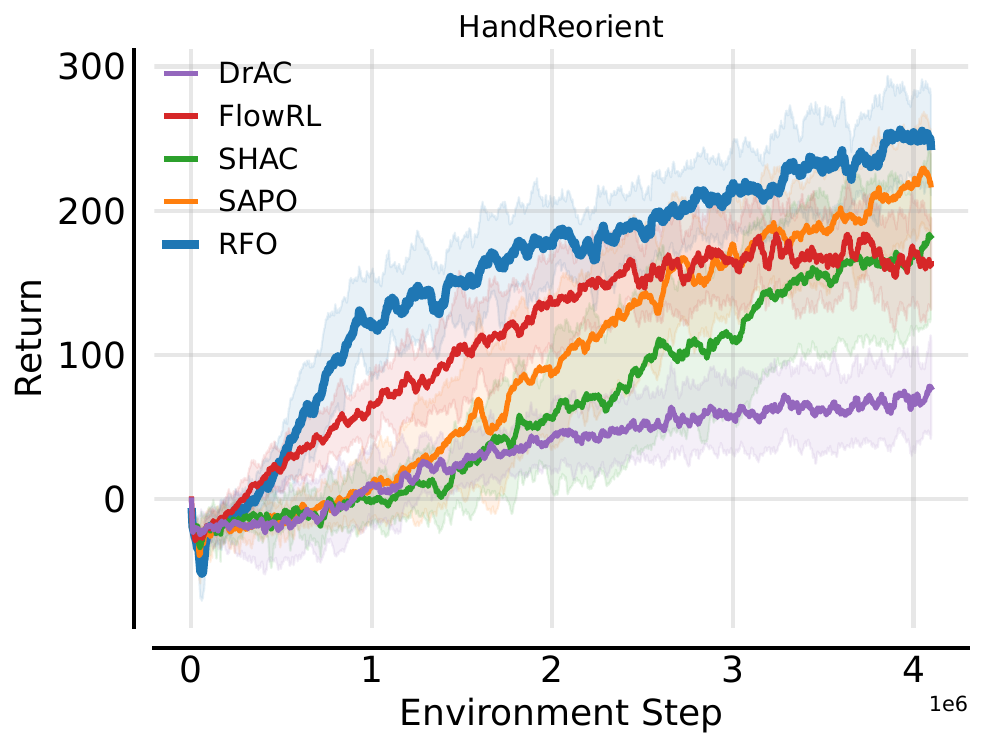}
    \hfill
    \includegraphics[width=0.24\textwidth]{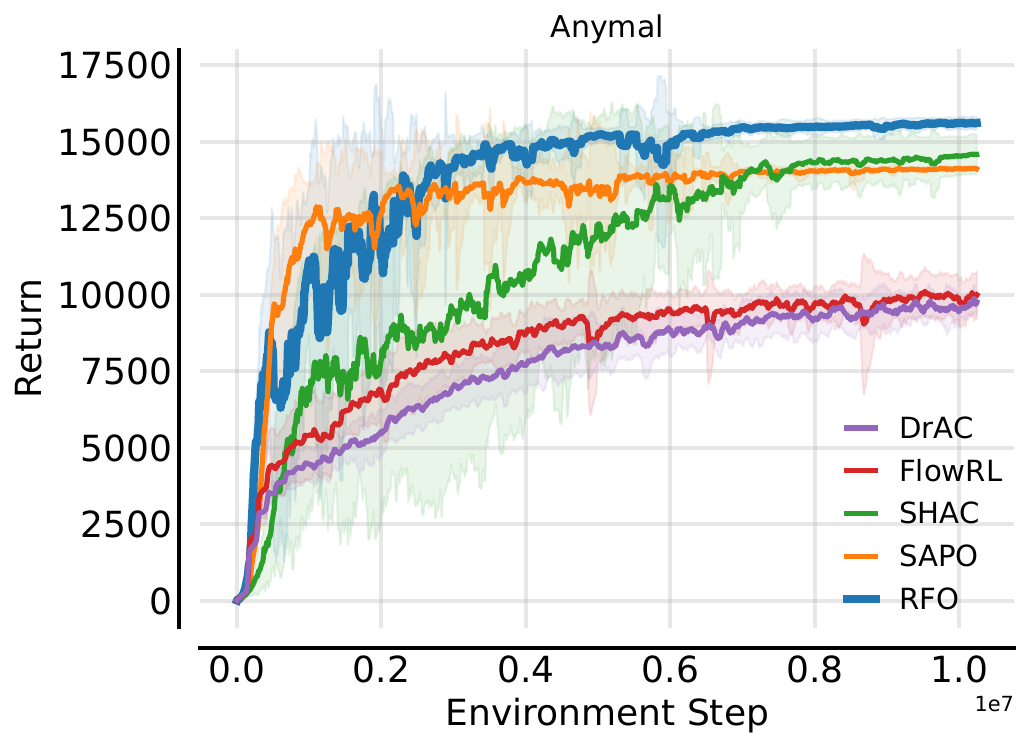}
    
    \vspace{2mm}

    % --- 第二行：三张图居中 ---
    \makebox[\textwidth][c]{%
        \includegraphics[width=0.24\textwidth]{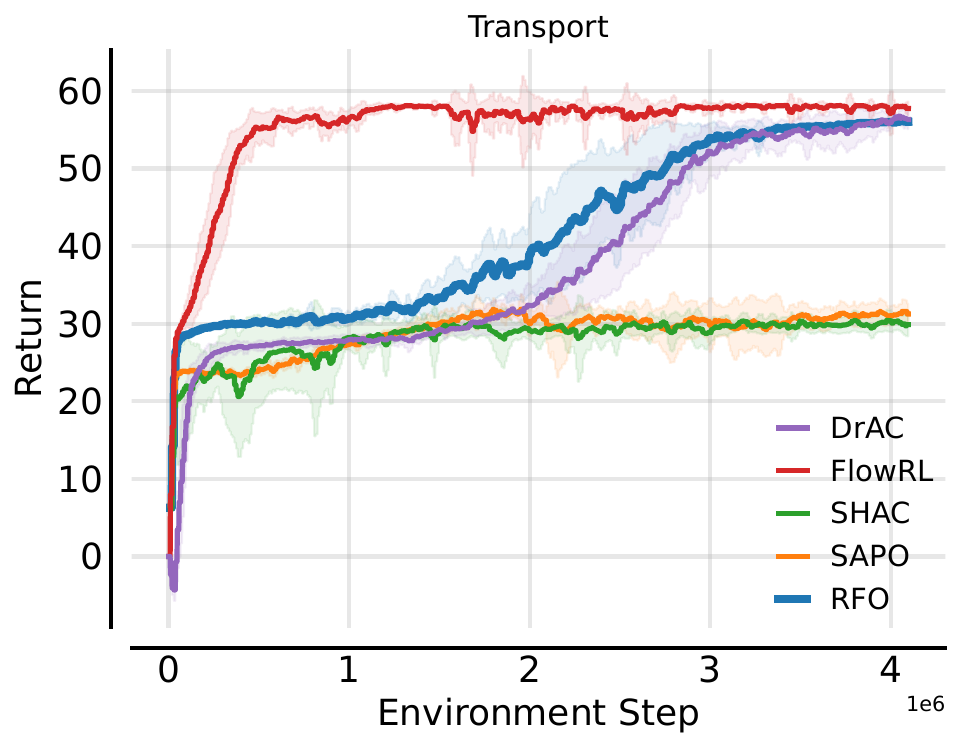}
        \quad
        \includegraphics[width=0.24\textwidth]{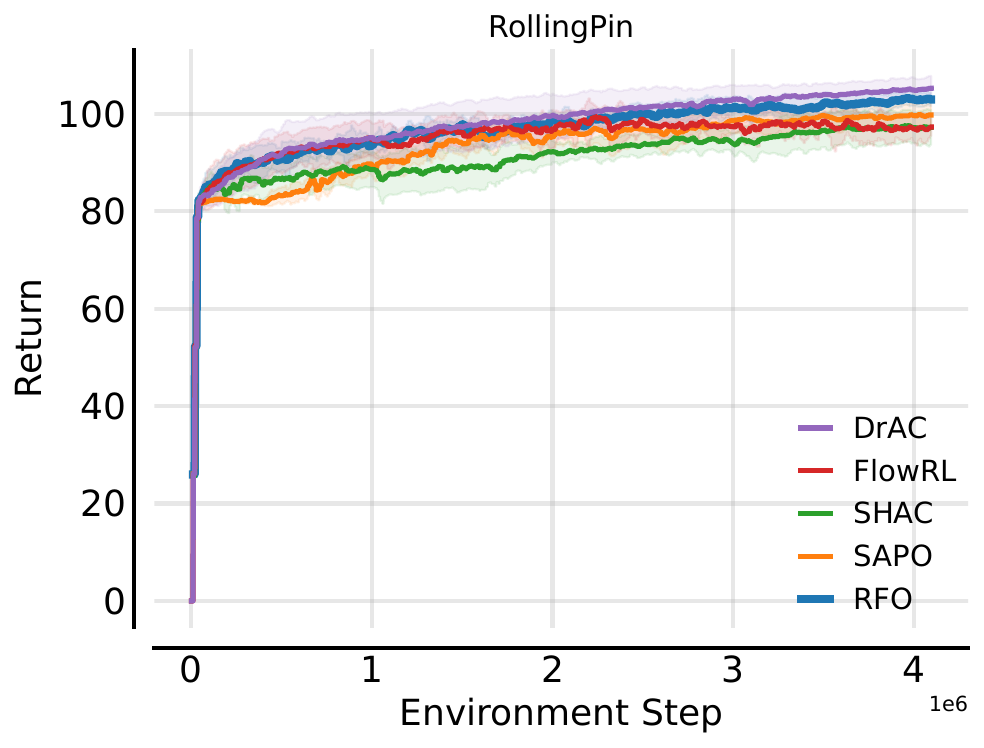}
        \quad
        \includegraphics[width=0.24\textwidth]{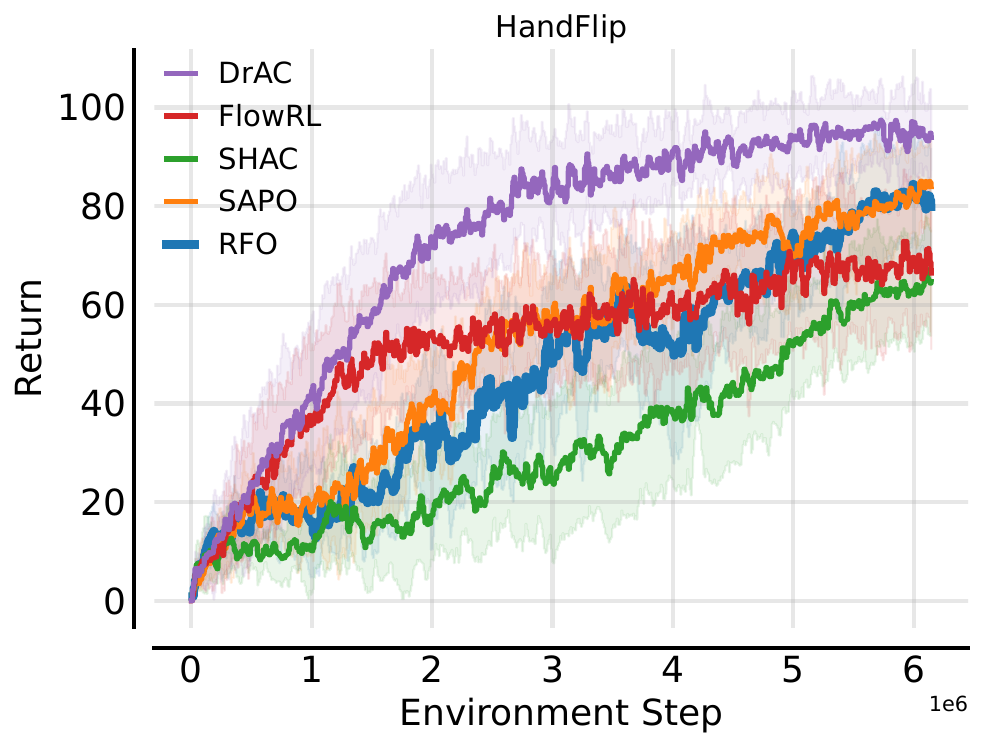}
    }

    \caption{Training curves across seven tasks. Solid lines represent the mean return, while shaded regions indicate the standard deviation. Curves are smoothed using a 100-episode moving average.}
    \label{fig:main_training_curves}

\end{figure*}

\begin{table*}
\centering
\caption{Final performance evaluation after training. Results are reported as mean $\pm$ standard deviation across all random seeds. 
Each seed is evaluated over 128 episodes. Normalized score $= \frac{\text{Mean Reward}}{\text{SHAC Mean Reward}}$.}
\label{table:Final_Eval}

\subcaption*{(a) Raw Final Evaluation Rewards}
\vspace{0.3em}
\begin{small}
\begin{tabular}{lccccc}
    \toprule
    \textbf{Task} & \textbf{DrAC} & \textbf{FlowRL} & \textbf{SHAC} & \textbf{SAPO} & \textbf{RFO (Ours)} \\
    \midrule
    Soft Jumper 
    & 83.56 $\pm$ 67.60 & -75.71 $\pm$ 139.54 & 1148.87 $\pm$ 233.84 & 1597.99 $\pm$ 653.87 & \textbf{3023.96 $\pm$ 603.61} \\
    Ant 
    & 4846.41 $\pm$ 1412.76 & 4083.45 $\pm$ 1082.79 & 3143.08 $\pm$ 569.50 & 4865.36 $\pm$ 363.77 & \textbf{5677.88 $\pm$ 964.49} \\
    Hand Reorient 
    & 73.74 $\pm$ 34.22 & 135.57 $\pm$ 39.34 & 174.63 $\pm$ 57.54 & 213.44 $\pm$ 33.95 & \textbf{258.84 $\pm$ 21.80} \\
    ANYmal 
    & 9554.46 $\pm$ 928.33 & 9717.58 $\pm$ 1619.35 & 14568.97 $\pm$ 652.72 & 14095.90 $\pm$ 82.02 & \textbf{15622.76 $\pm$ 200.87} \\
    Transport 
    & 57.07 $\pm$ 0.85 & \textbf{58.04 $\pm$ 0.19} & 30.00 $\pm$ 1.43 & 31.35 $\pm$ 1.24 & 56.00 $\pm$ 0.28 \\
    Rolling Pin 
    & \textbf{104.71 $\pm$ 2.00} & 92.49 $\pm$ 2.88 & 97.29 $\pm$ 3.44 & 99.79 $\pm$ 2.42 & 103.11 $\pm$ 1.26 \\
    Hand Flip 
    & \textbf{89.82 $\pm$ 4.59} & 54.80 $\pm$ 18.55 & 65.57 $\pm$ 11.75 & 83.00 $\pm$ 7.49 & 82.65 $\pm$ 11.65 \\
    \bottomrule
\end{tabular}

\vspace{1em}
\subcaption*{(b) SHAC-normalized Final Evaluation Rewards}
\vspace{0.3em}
\begin{tabular}{lccccc}
    \toprule
    \textbf{Task} & \textbf{DrAC} & \textbf{FlowRL} & \textbf{SHAC} & \textbf{SAPO} & \textbf{RFO (Ours)} \\
    \midrule
    Soft Jumper   & 0.07 $\pm$ 0.06 & -0.07 $\pm$ 0.12 & 1.00 $\pm$ 0.20 & 1.39 $\pm$ 0.57 & \textbf{2.63 $\pm$ 0.53} \\
    Ant           & 1.54 $\pm$ 0.45 & 1.30 $\pm$ 0.34  & 1.00 $\pm$ 0.18 & 1.55 $\pm$ 0.12 & \textbf{1.81 $\pm$ 0.31} \\
    Hand Reorient & 0.42 $\pm$ 0.20 & 0.78 $\pm$ 0.23  & 1.00 $\pm$ 0.33 & 1.22 $\pm$ 0.19 & \textbf{1.48 $\pm$ 0.12} \\
    ANYmal        & 0.66 $\pm$ 0.06 & 0.67 $\pm$ 0.11  & 1.00 $\pm$ 0.04 & 0.97 $\pm$ 0.01 & \textbf{1.07 $\pm$ 0.01} \\
    Transport     & 1.90 $\pm$ 0.03 & \textbf{1.93 $\pm$ 0.01} & 1.00 $\pm$ 0.05 & 1.05 $\pm$ 0.04 & 1.87 $\pm$ 0.01 \\
    Rolling Pin   & \textbf{1.08 $\pm$ 0.02} & 0.95 $\pm$ 0.03 & 1.00 $\pm$ 0.04 & 1.03 $\pm$ 0.02 & 1.06 $\pm$ 0.01 \\
    Hand Flip     & \textbf{1.37 $\pm$ 0.07} & 0.84 $\pm$ 0.28 & 1.00 $\pm$ 0.18 & 1.27 $\pm$ 0.11 & 1.26 $\pm$ 0.18 \\
    \midrule
    \textbf{Average} & 1.01 & 0.91 & 1.00 & 1.21 & \textbf{1.60} \\
    \bottomrule
\end{tabular}
\end{small}
\end{table*}

\subsection{Action Chunking}
Action chunking has proven beneficial for improving the temporal consistency of robot action generation~\cite{black2025realtime}, a critical factor in robotic applications. Therefore, we provide an extension to RFO that incorporates action chunking. In this variant, the flow policy predicts an action chunk of size $C$ based on the current observation, encompassing both current and future actions. Subsequently, the agent executes these actions for $C$ steps before replanning based on the new observation.

\section{Experiments}
We conduct experiments to answer the following questions: \textbf{(i)} How does RFO's performance compare to RPG-based approaches with Gaussian policies? \textbf{(ii)} How does RFO's performance compare to other RL algorithms for flow or diffusion-based approaches? \textbf{(iii)} Are the two regularization terms critical to RFO's performance? \textbf{(iv)} How does RFO work with the action chunking mechanism?
\subsection{Experimental Setup}
We evaluate our method on a comprehensive set of locomotion and manipulation tasks, utilizing either state or visual inputs within the differentiable simulators Rewarped~\cite{SAPO} and DFlex~\cite{SHAC,AHAC}.

\paragraph{Tasks.} Our experiments span diverse rigid and soft-body dynamics, categorized into:
\begin{itemize}
    \item Locomotion tasks: Ant, ANYmal~\cite{anymal} (rigid quadruped), and Soft Jumper (visual soft body).
    \item Manipulation tasks: Hand Reorient (cube rotation), Rolling Pin (dough flattening), Hand Flip (object flipping), and Transport (liquid transport).
\end{itemize}
Visual inputs are used for Soft Jumper and the last three manipulation tasks.

\paragraph{Baselines.} We compare RFO against strong baselines from two categories:
\begin{itemize}
    \item RPG Baselines: SAPO~\cite{SAPO} (SOTA RPG-based method with maximum entropy) and SHAC~\cite{SHAC} (short-horizon RPG).
    \item Flow/Diffusion RL Baselines: DrAC~\cite{DRAC} (a SOTA diffusion policy) and FlowRL~\cite{FuchunFlow} (a SOTA flow RL method).
\end{itemize}
All tasks are evaluated using at least 10 random seeds for each algorithm.

\begin{figure*} [t]

    % --- 第二行：三张图居中 ---
    \makebox[\textwidth][c]{%
        \includegraphics[width=0.33\textwidth]{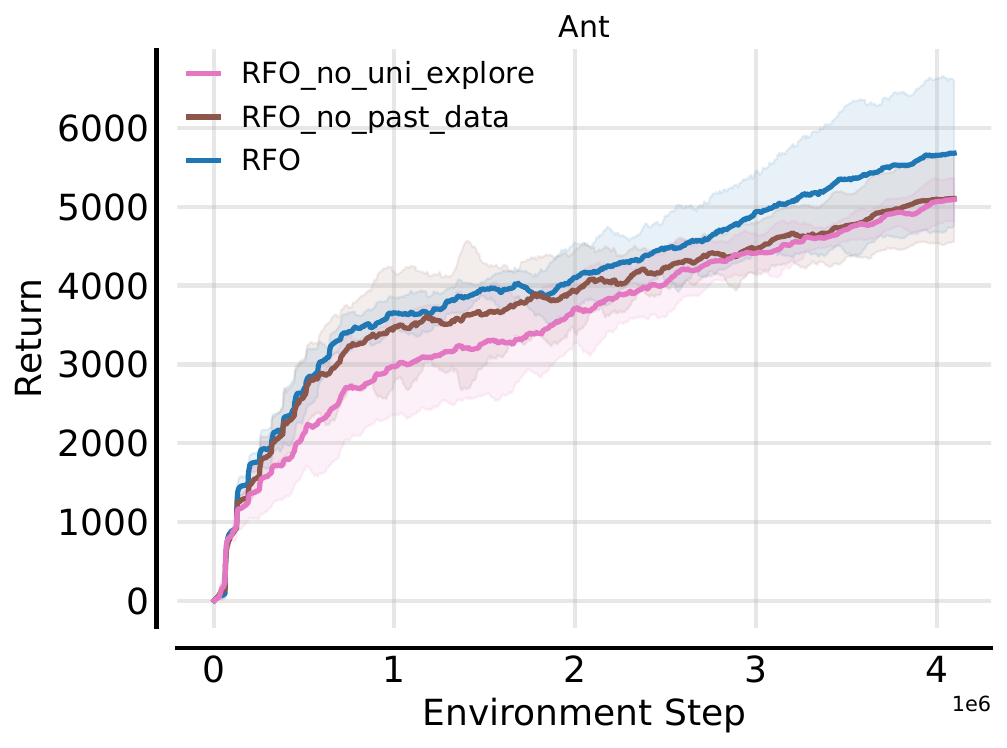}
        \quad
        \includegraphics[width=0.33\textwidth]{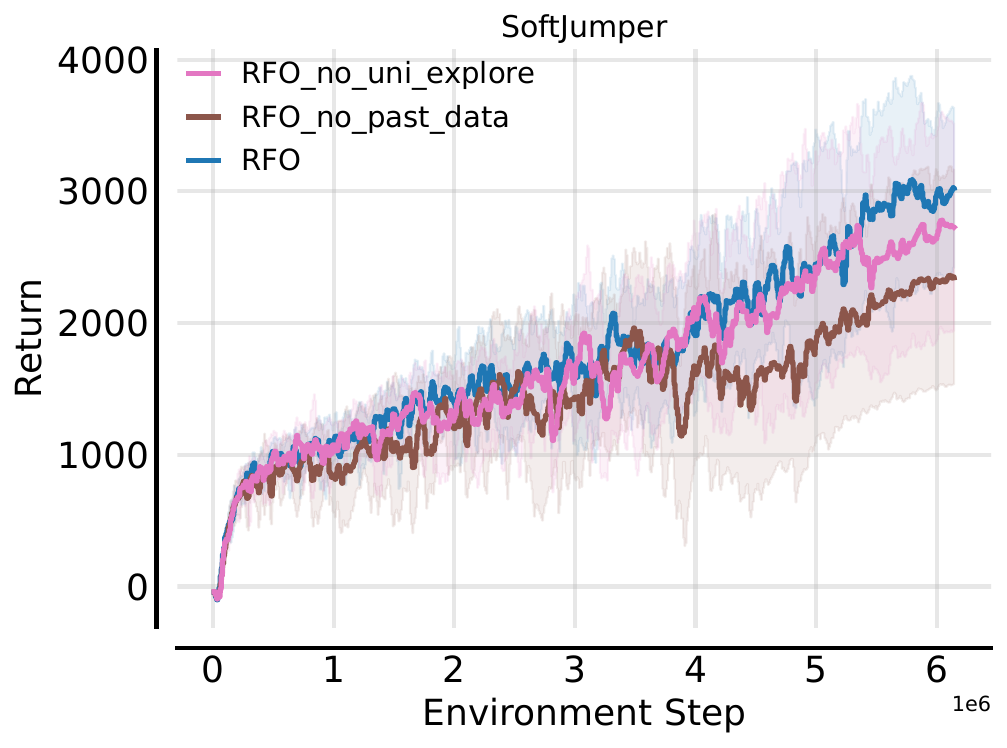}
        \quad
        \includegraphics[width=0.33\textwidth]{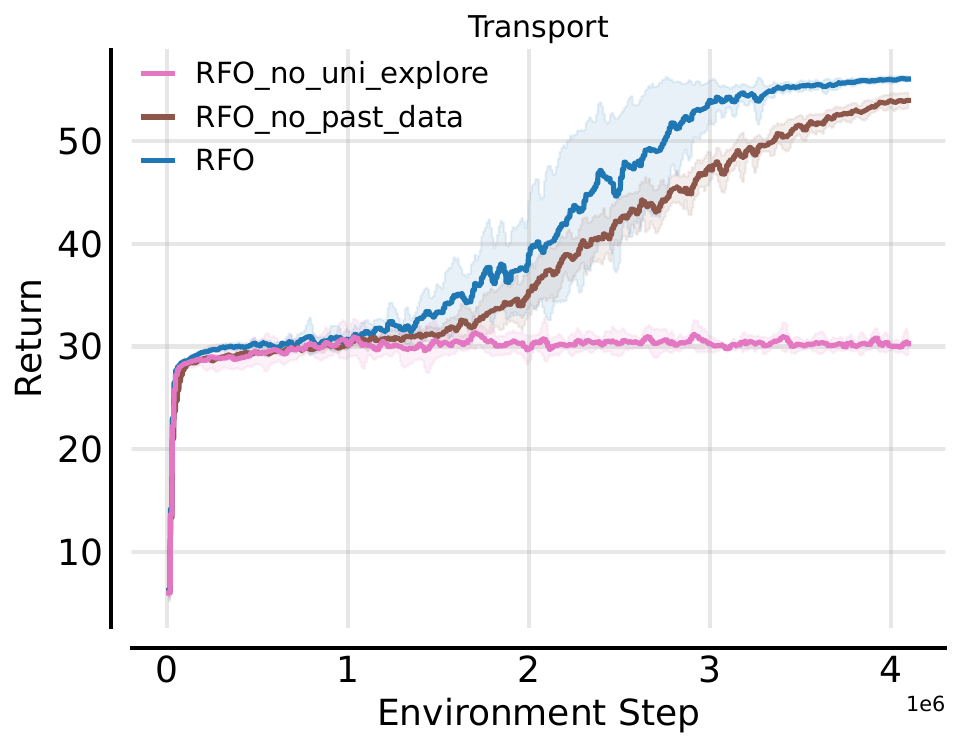}
    }

    \caption{Ablation study for the effectiveness of past data CFM regularization and uniform exploration CFM regularization. The results show that the two proposed regularization terms are critical for RFO's performance.}
    \label{fig:ablation_regularization}

\end{figure*}

\subsection{Main Experiment Results}
We show the training curves in Figure~\ref{fig:main_training_curves} and final performance evaluation results in Table~\ref{table:Final_Eval}.

In the Soft Jumper, Ant, ANYmal, and Hand Reorient tasks, RFO outperforms all baselines by substantial margins. Most notably, in the Soft Jumper task, RFO achieves a final reward nearly $2\times$ that of the best baseline, underscoring its capability in high-dimensional control settings (action space $\mathbb{R}^{222}$). This superiority extends to rigid-body dynamics, where RFO consistently dominates strong baselines, securing performance gains of approximately 21\% on Hand Reorient and 17\% on Ant, while boosting the reward by over $1000$ points on the ANYmal task. Furthermore, RFO demonstrates remarkable robustness on the Rolling Pin and Transport tasks, achieving performance comparable to the state-of-the-art with marginal gaps of only 1.5\% and 3.5\%, respectively. As shown in the aggregate performance analysis (Table~\ref{table:Final_Eval} (b)), which normalizes scores against the SHAC baseline, RFO delivers superior overall performance across the diverse task suite.

From the RPG perspective, RFO consistently improves upon SHAC and SAPO, validating the efficacy of integrating flow policies into the RPG framework. Compared to Flow and Diffusion baselines, RFO demonstrates a clear advantage by leveraging analytical gradients from RPG. This becomes particularly evident in the Soft Jumper, ANYmal, and Hand Reorient tasks, where DrAC and FlowRL struggle, while RFO demonstrates superior performance. Additionally, DrAC requires $20$ denoising steps for action sampling, which not only slows down inference but also significantly increases training time compared to RFO, since it needs to backpropagate through all $20$ action generation steps. Furthermore, despite being an on-policy algorithm, RFO achieves sample efficiency comparable to off-policy flow-based methods thanks to RPG, while avoiding the need to approximate intractable action log-likelihoods.

\subsection{Ablation Study} \label{sec:main_ablation}
In this section, we conduct ablation experiments to answer the following questions: \textbf{(i)} Are the two proposed regularization terms, past data CFM regularization and uniform exploration CFM regularization, critical for RFO's performance? \textbf{(ii)} How does RFO perform under a range of hyperparameters, including the weights for the two CFM objectives and the number of flow integration steps?

To answer the first question, we conduct experiments on the Ant, Soft Jumper, and Transport tasks. We compare the full RFO algorithm against two variants: RFO without past data CFM regularization and RFO without uniform exploration CFM regularization. The training curves are presented in Figure~\ref{fig:ablation_regularization}. 

Without past data CFM regularization, RFO's performance significantly degrades across all three tasks, suggesting that this term is crucial for both training stability and final performance. Similarly, removing the uniform exploration regularization leads to performance drops across all tasks. Most notably, on the Transport task, omitting uniform exploration regularization causes RFO's performance to degrade to the levels of SHAC and SAPO, clearly highlighting the effectiveness of the uniform random exploration strategy.

To answer the second question, we conduct experiments on the Soft Jumper task. We evaluate RFO with different combinations of $c_{\text{past}}$ and $c_{\text{uni}}$, which control the weights of the two regularization terms. We also evaluate RFO with varying numbers of flow Euler integration steps. The training curves are provided in Appendix~\ref{appendix:hypersens}, demonstrating that RFO exhibits strong robustness across a wide range of hyperparameters.

\subsection{Action Chunking Results}
Figure~\ref{fig:appendix_action_chunk_results} in Appendix~\ref{appendix:action_chunking} illustrates the training curves for RFO with action chunking across the evaluated tasks. This result is particularly encouraging given the inherent challenge: the flow policy must predict action sequences for future time steps based solely on the current observation, without the guidance of expert data used in imitation learning. Such optimization is harder, as the policy must generate valid future actions without access to the corresponding intermediate states. Consequently, this setup suggests that integrating offline-to-online procedures could be a promising direction to further enhance RPG-based flow policies.

\section{Conclusion}
In this work, we introduce Reparameterization Flow Policy Optimization, a novel framework that bridges the gap between flow-based generative policies and the high sample efficiency of Reparameterization Policy Gradients. By integrating tailored regularization terms for training stability and exploration, RFO achieves strong performance. Empirical evaluations across diverse rigid and soft-body control tasks demonstrate that RFO achieves SOTA performance. We further explored an action-chunking RFO variant. Future directions include extending RFO to offline-to-online reinforcement learning settings.

\bibliography{plain}
\bibliographystyle{plain}

\appendix

%barrier to its application is training instability, where high-variance gradients can destabilize the learning process. To address this, we draw      \link{Extended version}{https://aaai.org/example/extended-version}

\newpage
\section{Task Details} \label{appendix:task_details}
We conduct experiments with two differentiable physics simulators: DFlex~\cite{SHAC,AHAC} from NVIDIA and Rewarped~\cite{SAPO}. More specifically, we use AHAC's version of the DFlex simulator and Rewarped version 1.3.0 from \url{https://github.com/rewarped/rewarped}.

Here, we provide details of the tasks used in the main experiments.

\subsection{Ant}

Ant is a classical locomotion task using the Rewarped simulator~\cite{SAPO}, where the goal is to maximize forward velocity. Ant uses state inputs. The state space is $R^{37}$ and the action space is $R^8$, including joint torques~\cite{SAPO}.

\subsection{Hand Reorient}
Hand Reorient is a task from Rewarped~\cite{SAPO}, where the goal is to reorient a cube with an Allegro dexterous hand, using state inputs. The state space is $R^{72}$ and the action space is $R^{16}$, including joint torques~\cite{SAPO}.

\subsection{Rolling Pin}
Rolling Pin is a manipulation task from Rewarped~\cite{SAPO}, where the goal is to flatten dough with a rolling pin, using pixel inputs. The observation space consists of $\left [R^{250 \times3}, R^3, R^3 \right]$, and the action space is $R^3$, including relative position and orientation~\cite{SAPO}.

\subsection{Soft Jumper}
Soft Jumper is a soft-body locomotion task from Rewarped~\cite{SAPO} with pixel inputs. The observation space consists of $\left [R^{204 \times3}, R^3, R^3, R^{222}\right]$, and the action space is $R^{222}$, including tetrahedral activations~\cite{SAPO}.

\subsection{Hand Flip}
Hand Flip is a soft-object manipulation task from Rewarped~\cite{SAPO} using a Shadow Hand and pixel inputs. The observation space consists of $\left [R^{250 \times3}, R^3, R^{24}\right]$, and the action space is $R^{24}$, including relative position and orientation~\cite{SAPO}.

\subsection{Transport}
Transport is a soft-object manipulation task from Rewarped~\cite{SAPO} with pixel inputs. The observation space consists of $\left [R^{250 \times3}, R^3, R^3, R^3\right]$, and the action space is $R^3$, including relative position~\cite{SAPO}.

\subsection{ANYmal}
ANYmal is a locomotion task from DFlex~\cite{AHAC} with state inputs, where the goal is to maximize the forward velocity of a quadruped robot~\cite{anymal}. The state space is $R^{49}$ and the action space is $R^{12}$.

% \newpage

\section{Baselines, Implementation Details and Hyperparameter} \label{appendix:impl_hyper}
\subsection{Baselines and Hyperparameters}

We benchmark RFO against four baselines: SHAC~\cite{SHAC}, SAPO~\cite{SAPO}, DrAC~\cite{DRAC}, and FlowRL~\cite{FuchunFlow}. For SAPO, we utilize the official repository\footnote{\url{https://github.com/etaoxing/mineral}}, from which we also adopt the improved SHAC implementation as it demonstrates improved and stable performance. For DrAC (Diffusion Policy) and FlowRL (Flow Policy), we adapt their official implementations to support parallelized environments and incorporate visual encoders for pixel-based inputs. We use their default sampler and integration steps for action sampling.

We tune the hyperparameters of DrAC and FlowRL individually for each task, finding that the Update-to-Data (UTD) ratio affects their performance significantly (FlowRL defaults to $0.5$, as in their official repository). Note that tuning DrAC with UTD ratios $\ge 1$ is computationally prohibitive, as training would require weeks to complete. %Hence, we tune DrAC and FlowRL's UTD ratios from $\left[0.125,0.25,0.5 \right]$ on each task. 
Furthermore, we observe that a higher UTD ratio does not necessarily translate to better performance. We align the neural network sizes across all algorithms. For visual input tasks, we use DP3 Point Net with layers [64, 128, 256] and 64-dimension output, average pooling for all algorithms. Shared hyperparameters are detailed in Table~\ref{tab:shared_params}.

\begin{table*}[!ht]
\centering
\caption{Common hyperparameters for all algorithms.}
\label{tab:shared_params}
%\makebox[\linewidth][c]{
\setlength{\tabcolsep}{0.2em}
\small
\begin{tabular}{lcccccc}
    \toprule
     & \textit{shared} & RFO & SHAC & SAPO & DrAC &FlowRL\\
    \midrule
    Horizon $H$ & 32 \\
    Epochs for critics $L$ & & 16 &16&16&see UTD ratio & see UTD ratio  \\
    Epochs for actors $M$ & & 1 & 1 & 1 & see UTD ratio & see UTD ratio\\
    Discount $\gamma$ & $0.99$ \\
    TD/GAE $\lambda$ & 0.95& &  &  & N.A. & N.A.\\
    Actor MLP & $(400,200,100)$ &  \\
    Critic MLP & $(400,200,100)$ &  \\    
    Actor $\eta$ & & $2e-3$ & $2e-3$ & $2e-3$ & $3e-4$ & $3e-4$\\
    Critic $\eta$ & $5e-4$& & &  & $3e-4$ & $3e-4$  \\

    Learning rate schedule & - & linear &linear & linear & N.A. &N.A. \\
    Optim type & AdamW  \\
    Optim $(\beta_1, \beta_2)$ &$(0.7, 0.95)$  &$(0.9, 0.999)$ &$(0.9, 0.999)$ & $(0.9, 0.999)$ &  &\\
    %Grad clip & $0.5$ &&&& N.A.&N.A.\\
    Norm type & LayerNorm  \\
    Activation type & SiLU  \\
    \bottomrule
\end{tabular}

\end{table*}

\begin{table*}[!ht]
\centering
\caption{The number of parallel environments used for each environment. These values are kept the same as in the official implementations: we follow the AHAC repository (\url{https://github.com/imgeorgiev/DiffRL}) for the DFlex tasks and the Rewarped repository.}
\label{tab:Num_envs}
\begin{tabular}{lccccccc}
    \toprule
     & Ant & Transport  & Hand Flip&Rolling Pin&Soft Jumper & Anymal & Hand Reorient \\
    \midrule
    Num Envs &64&32   &32&32&32&128 & 64 \\
    \bottomrule
\end{tabular}

\end{table*}

\begin{table*}[!ht]
\centering
\caption{UTD ratios for FlowRL and DrAC.}
\label{tab:UTD}
\begin{tabular}{lccccccc}
    \toprule
     & Ant & Transport  & Hand Flip&Rolling Pin&Soft Jumper & Anymal & Hand Reorient \\
    \midrule
    DrAC &0.5 & 0.125  &0.5&0.25&0.125& 0.25 & 0.125 \\
    FlowRL &0.125& 0.125  &0.25&0.5&0.5&0.25 & 0.25 \\
    \bottomrule
\end{tabular}
\end{table*}

\subsection{Implementation Details of RFO}
Since the flow velocity field can generate unbounded actions, we apply a tanh squashing function to the output. We utilize the pre-tanh values as targets for both past data CFM regularization and uniform CFM regularization (although we found that using tanh-transformed actions as targets is also effective). Additionally, we observe that excessively large pre-tanh targets can cause issues for the CFM loss; hence, we apply clamping to bound them. Detailed unique hyperparameters for RFO are shown in Table~\ref{tab:rfo_hyperparams}.

\begin{table*}[!ht]
\centering
\caption{RFO's unique hyperparameters.}
\label{tab:rfo_hyperparams}
\begin{tabular}{lccccccc}
    \toprule
     & Ant & Transport  & Hand Flip&Rolling Pin&Soft Jumper & Anymal & Hand Reorient  \\
    \midrule
    $c_{\text{past}}$ &0.2 & 0.1  &5e-4&5e-3&0.4&0.1  &0.2  \\
    $c_{\text{uni}}$  &0.2 & 0.1  &0.2&5e-3&0.4&0.1  & 0.01 \\

    \bottomrule
\end{tabular}

\end{table*}

\begin{figure*}[t]

    % --- 第二行：三张图居中 ---
    \makebox[\textwidth][c]{%
        \includegraphics[width=0.4\textwidth]{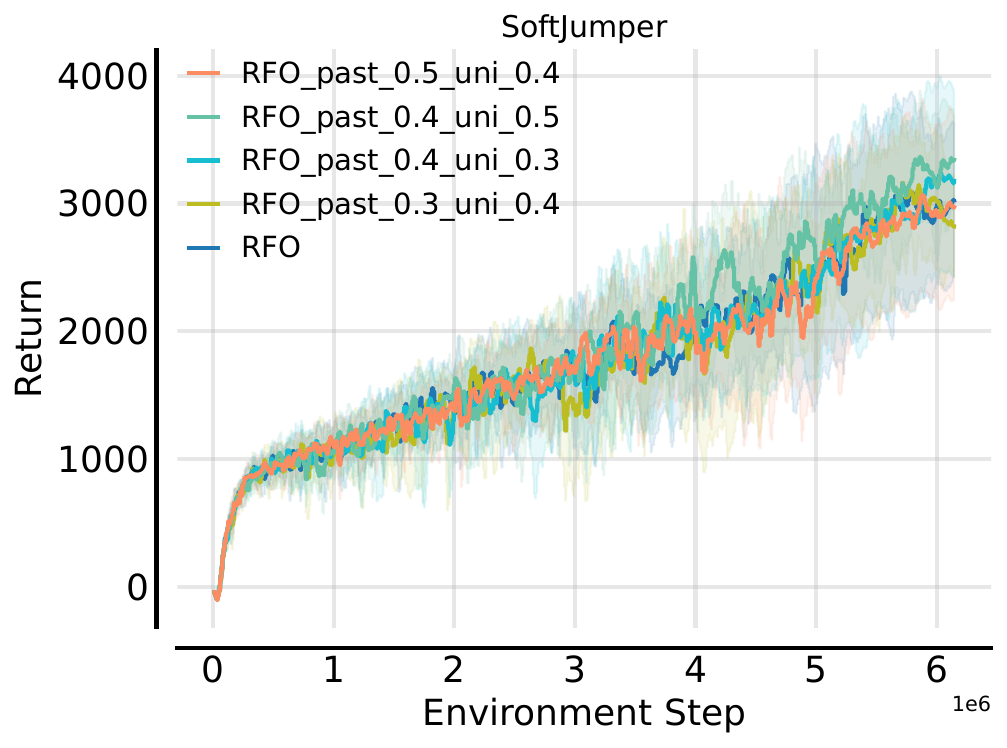}
        \quad
        \includegraphics[width=0.4\textwidth]{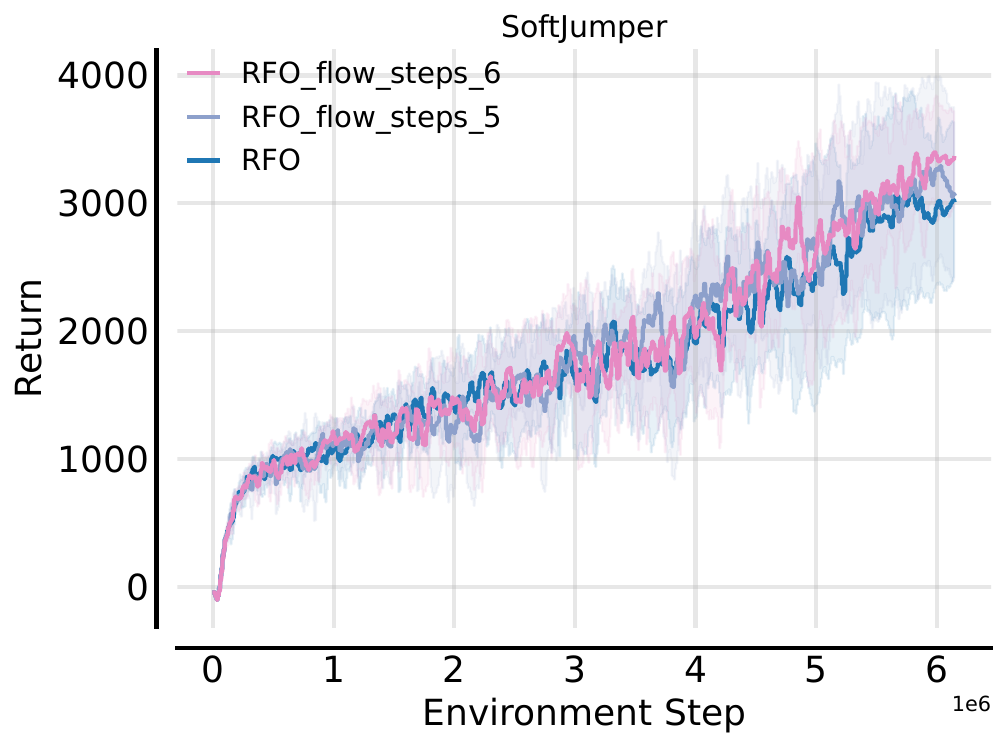}

    }

    \caption{(a) Ablation study on different weight combinations for past data CFM regularization and uniform exploration CFM regularization. (b) Ablation study on the number of flow integration steps.}
    \label{fig:appendix_ablation_hyperparameters}

\end{figure*}

\section{Sensitivity to Hyperparameters} \label{appendix:hypersens}
In this section, we conduct ablation experiments on: (i) different weight combinations for past data CFM regularization and uniform exploration CFM regularization, and (ii) the number of flow integration steps (RFO uses 4 flow steps across all tasks by default). As shown in Figure~\ref{fig:appendix_ablation_hyperparameters}, RFO demonstrates robust performance across a wide range of hyperparameters.

\section{RFO with Action Chunking Results} \label{appendix:action_chunking}
\begin{figure*}[!ht]
    \centering
    % --- 第一行 ---
    \includegraphics[width=0.32\textwidth]{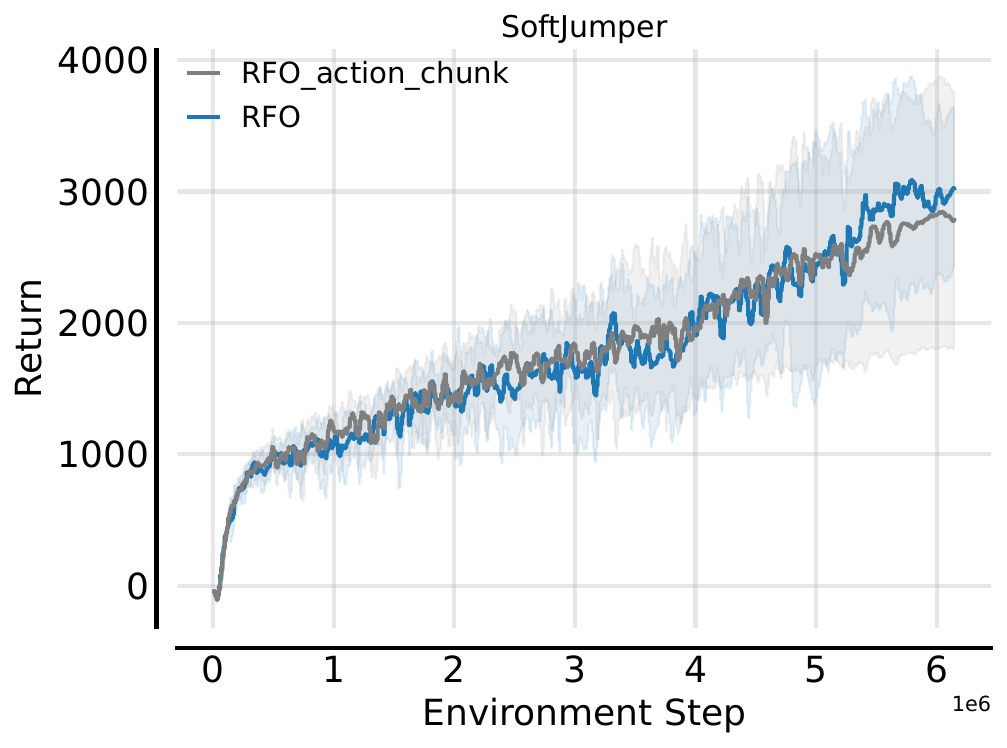}
    \hfill % 自动填充间距
    \includegraphics[width=0.32\textwidth]{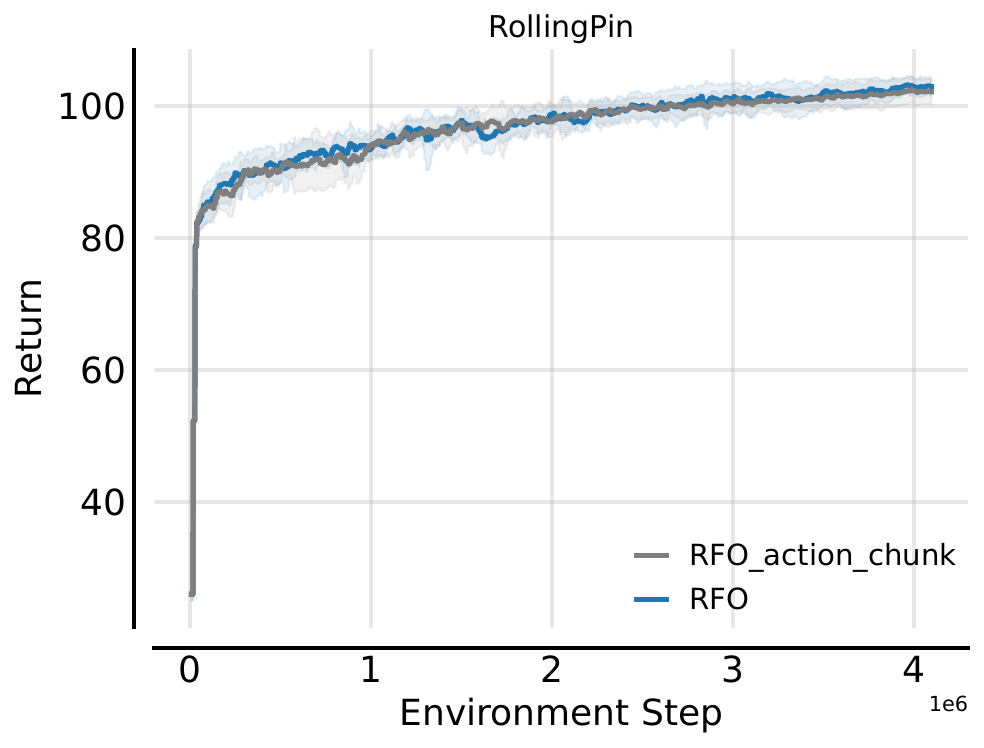}
    \hfill
    \includegraphics[width=0.32\textwidth]{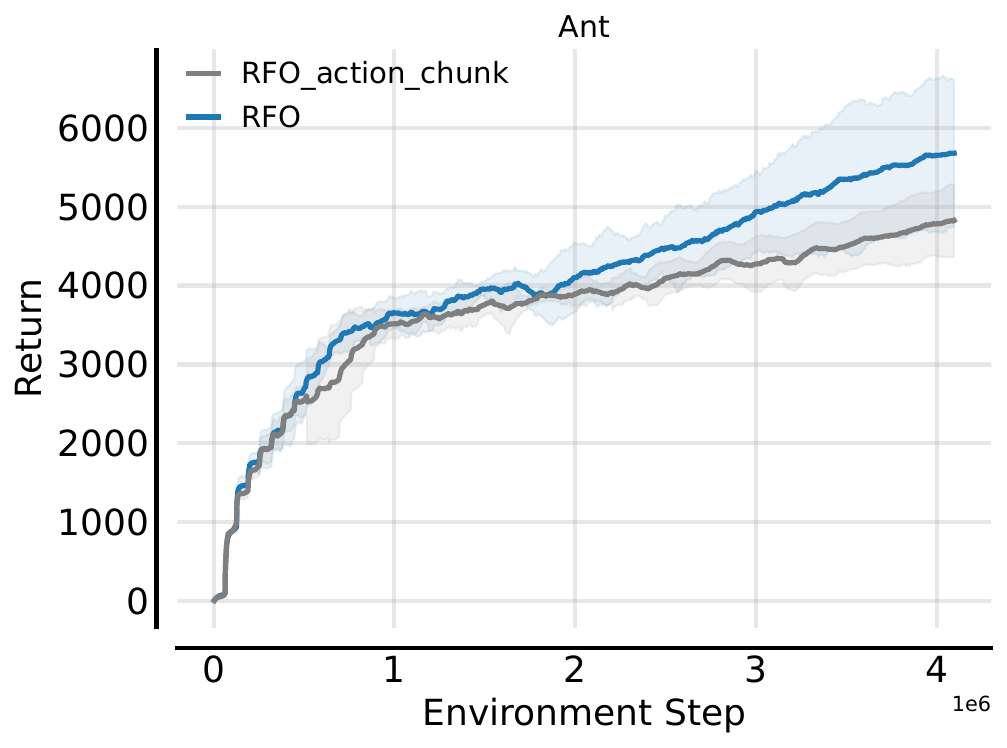}
    % --- 第二行 (使用第一行的图作为 Placeholder) ---
    % 请把下面的路径换成你真正需要的图片
    \includegraphics[width=0.32\textwidth]{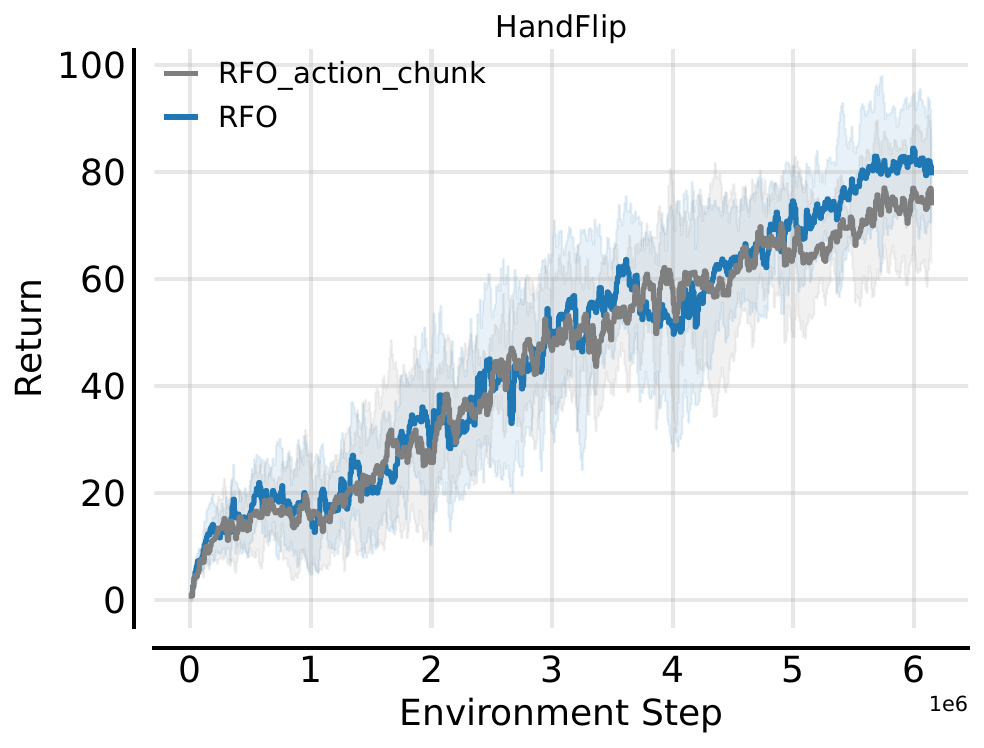} 
    \hfill
    \includegraphics[width=0.32\textwidth]{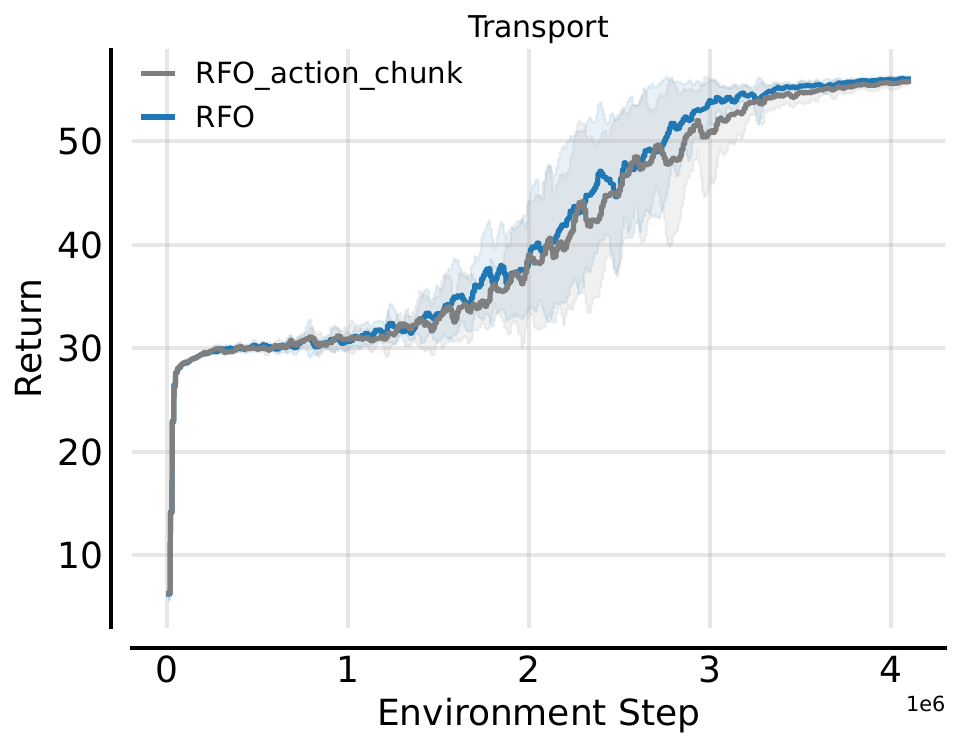}
    \hfill
    \includegraphics[width=0.32\textwidth]{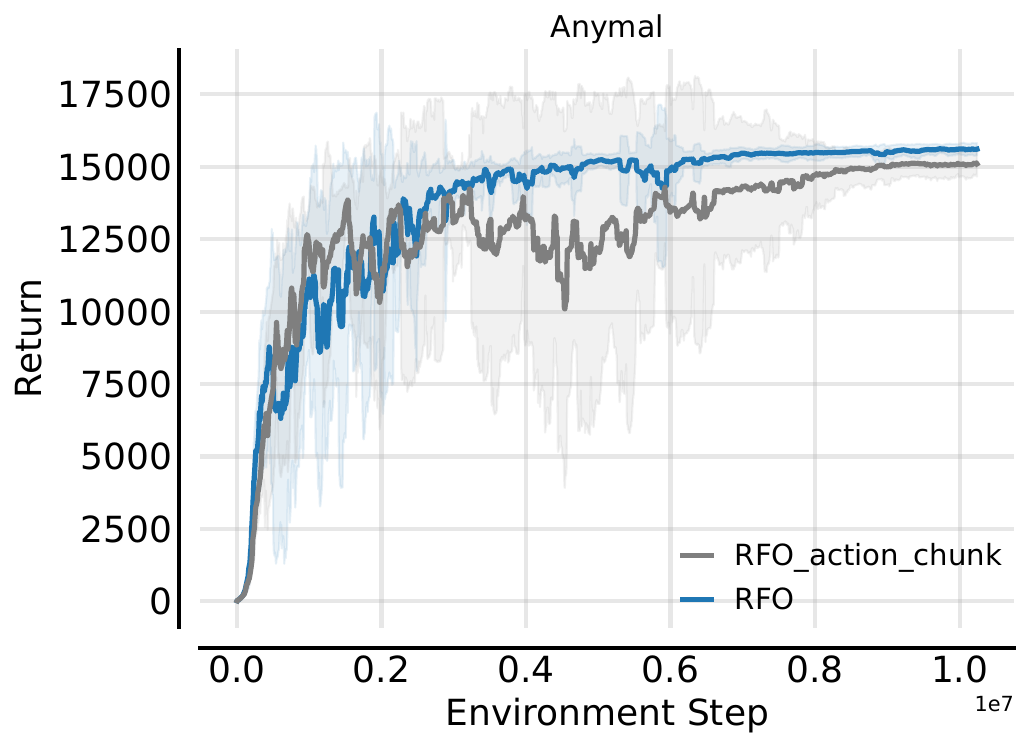}

    \caption{Training curves of RFO with the action chunking mechanism.}
    \label{fig:appendix_action_chunk_results}
\end{figure*}

Figure~\ref{fig:appendix_action_chunk_results} illustrates the training curves for RFO augmented with the action chunking mechanism. We observe that RFO with action chunking generally maintains strong performance, largely matching that of the standard RFO baseline (without chunking) across the majority of evaluated environments. Although a minor performance degradation is observed in certain task, the results remain competitive. This is expected, as the flow policy is required to predict a sequence of actions simultaneously, which increases the complexity of policy optimization.

\section{Details of KL Divergence Estimation} \label{appendix:KL}
In this section, we detail the approximation of the KL divergence between consecutive policy updates on the Soft Jumper task. Following the methodology in FPO~\cite{FLowmatchingPolicyGradients}, for a given state-action pair $(s, a)$ and flow policies $\pi_{\theta}$ and $\pi_{\theta'}$, we can approximate the likelihood ratio $\frac{\pi_{\theta}(a|s)}{\pi_{\theta'}(a|s)}$ using the difference in their Conditional Flow Matching (CFM) losses. Specifically, this ratio is approximated as $\exp(L_{\theta'}^{\text{CFM}}(a|s) - L_{\theta}^{\text{CFM}}(a|s))$. Leveraging this relationship, we sample actions from the old/behavior policy $\pi_{\theta}$ and compute the CFM losses for both $\pi_{\theta}$ and $\pi_{\theta'}$ using these samples as targets. Finally, we employ the K3 estimator~\cite{klblog} to estimate the KL divergence $\text{KL}(\pi_{\theta} \| \pi_{\theta'})$.

\section{Design Choice} \label{appendix:DesignChoice}
\begin{figure*}[!ht]
    \centering
    % --- 第一行 ---
    \includegraphics[width=0.4\textwidth]{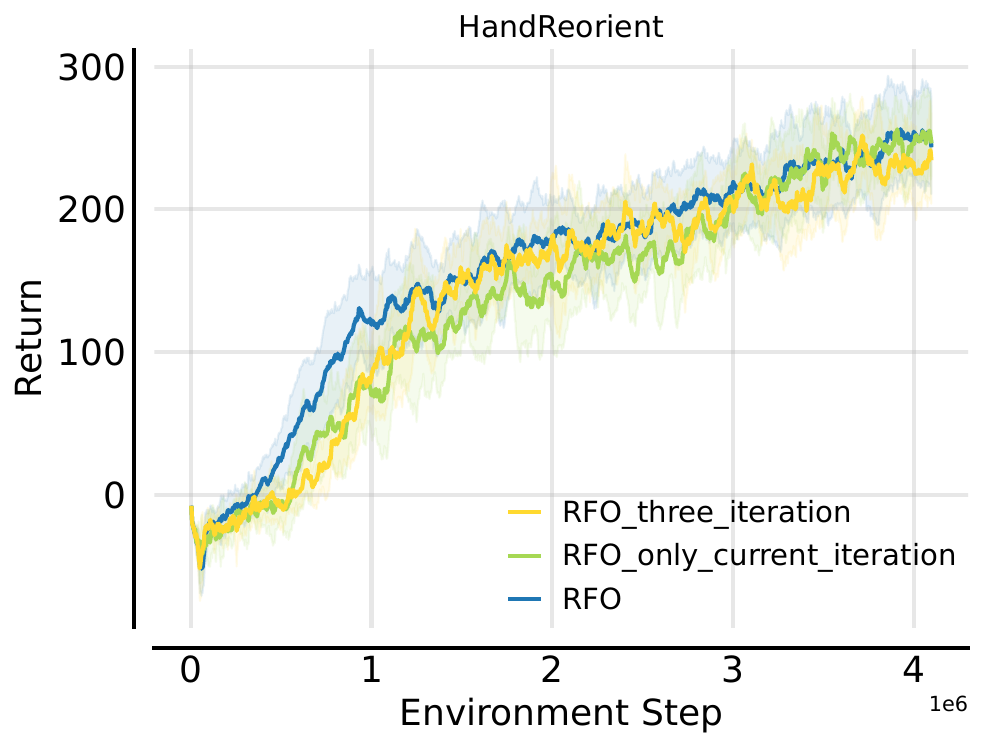}

    \caption{Comparison of $\mathcal{D}_{\text{recent}}$ including data from: the current iteration only, the current + previous iterations, and three iterations of rollout data.}
    \label{fig:appendix_Design_Choice}
\end{figure*}

In this section, we justify our decision to include rollout states and actions from both the current and previous iterations in $\mathcal{D}_{\text{recent}}$. As shown in Figure~\ref{fig:appendix_Design_Choice}, relying solely on the current iteration results in significantly slower learning. Conversely, using data from three iterations leads to degraded performance and reduced learning speed. In contrast, we find that including state-action pairs from both the current and previous iterations yields the best results in terms of final performance and convergence speed.

%%%%%%%%%%%%%%%%%%%%%%%%%%%%%%%%%%%%%%%%%%%%%%%%%%%%%%%%%%%%%%%%%%%%%%%%%%%%%%%
%%%%%%%%%%%%%%%%%%%%%%%%%%%%%%%%%%%%%%%%%%%%%%%%%%%%%%%%%%%%%%%%%%%%%%%%%%%%%%%

\end{document}